\PassOptionsToPackage{table}{xcolor}%
\documentclass[10pt,letterpaper]{article}
\usepackage{alibaba-report}
\providecommand{\reportaffiliationfont}{}
\title{Seeing but Not Thinking: Routing Distraction in Multimodal Mixture-of-Experts}
\author{Haolei Xu\textsuperscript{*,1,2},
  Haiwen Hong\textsuperscript{*,$\dagger$,2},
  Hongxing Li\textsuperscript{1,2},
  Rui Zhou\textsuperscript{1},
  Yang Zhang\textsuperscript{1},\\
  Longtao Huang\textsuperscript{2},
  Hui Xue\textsuperscript{2},
  Yongliang Shen\textsuperscript{$\ddagger$,1},
  Weiming Lu\textsuperscript{$\ddagger$,1},
  Yueting Zhuang\textsuperscript{1}\\[6pt]
  {\reportaffiliationfont\mdseries\textsuperscript{1}Zhejiang University,\quad
   \textsuperscript{2}Yuvion Team, Alibaba Group}\\[4pt]
  {\normalfont\textsuperscript{*}Equal Contribution,\quad
   \textsuperscript{$\dagger$}Project Leader,\quad
   \textsuperscript{$\ddagger$}Corresponding Author}\\[3pt]
  {\normalfont\texttt{\{xuhaolei, syl, luwm\}@zju.edu.cn},\quad
   \texttt{honghaiwen.hhw@alibaba-inc.com}}}
\newcommand{\shortpapertitle}{Seeing but Not Thinking}
\renewcommand{\authorcontributions}{}
\drafttrue

\usepackage{bm}
\usepackage{makecell}
\usepackage{array}
\usepackage{wrapfig}%
\makeatletter
\patchcmd{\WF@putfigmaybe}
  {\setbox\WF@box\hbox{\lower\intextsep\box\WF@box}}
  {\global\advance\WF@size-\intextsep}
  {}{\PackageWarningNoLine{main}{Could not patch wrapfig top alignment}}
\makeatother
\usepackage{needspace}%
\usepackage{newunicodechar}
\newunicodechar{–}{\textendash}
\newunicodechar{—}{\textemdash}
\newunicodechar{‘}{\textquoteleft}
\newunicodechar{’}{\textquoteright}
\newunicodechar{“}{\textquotedblleft}
\newunicodechar{”}{\textquotedblright}
\definecolor{lightgreen}{RGB}{34, 139, 34}
\newcolumntype{C}[1]{>{\centering\arraybackslash}p{#1}}

\lstdefinestyle{promptstyle}{
    basicstyle=\ttfamily\small,
    breaklines=true,
    breakatwhitespace=false,
    breakindent=0pt,
    columns=fullflexible,
    keepspaces=true,
    showstringspaces=false,
    linewidth=\linewidth,
    frame=none,
    xleftmargin=0pt,
    framexleftmargin=0pt,
    numbers=none,
    backgroundcolor={},
    xrightmargin=0pt,
    upquote=true,
    aboveskip=0pt,
    belowskip=0pt
}
\newtcolorbox{promptbox}[1][]{breakable,enhanced,
  colback=black!3,colframe=black!15,boxrule=0.4pt,arc=1mm,
  left=3mm,right=3mm,top=1.5mm,bottom=1.5mm,before skip=8pt,after skip=8pt,
  fonttitle=\sffamily\bfseries\small,coltitle=black!80,colbacktitle=black!7,
  titlerule=0.4pt,toptitle=0.8mm,bottomtitle=0.8mm,#1}

\makeatletter
\titleformat{\paragraph}[runin]{\normalsize\bfseries}{}{0pt}{}[\@addpunct{.}]
\makeatother

\begin{document}
\maketitle
\begin{abstract}

Multimodal Mixture-of-Experts (MoE) models have achieved remarkable performance on vision-language tasks. However, we identify a puzzling phenomenon termed \textbf{\textit{Seeing but Not Thinking}}: models accurately perceive image content yet fail in subsequent reasoning, while correctly solving identical problems presented as pure text. Through systematic analysis, we first verify that cross-modal semantic sharing exists in MoE architectures, ruling out semantic alignment failure as the sole explanation. We then reveal that visual experts and domain experts exhibit layer-wise separation, with image inputs inducing significant routing divergence from text inputs in middle layers where domain experts concentrate. Based on these findings, we propose the \textbf{\textit{Routing Distraction}} hypothesis: when processing visual inputs, the routing mechanism fails to adequately activate task-relevant reasoning experts. To validate this hypothesis, we design a routing-guided intervention method that enhances domain expert activation. Experiments on three multimodal MoE models across six benchmarks demonstrate consistent improvements, with gains of up to 3.17\% on complex visual reasoning tasks. Our analysis further reveals that domain expert identification locates cognitive functions rather than sample-specific solutions, enabling effective transfer across tasks with different information structures.

\end{abstract}
\section{Introduction}

Mixture-of-Experts (MoE) architectures \cite{cai2025survey} have become the dominant paradigm for scaling large vision-language models \cite{lin2024moe,wang2025internvl3,kuang2025natural,tang2025survey,ding2026omnisift}, powering a wide range of downstream multimodal applications \cite{lu2026uir1,lu2025uis1}. By activating only a sparse subset of experts for each input, MoE models efficiently handle the intricate interactions between visual and textual information while maintaining computational tractability. However, beneath this success lies a puzzling phenomenon that challenges our fundamental understanding of how these models integrate perception and reasoning.

\begin{wrapfigure}{R}{0.45\linewidth}
  \centering
  \includegraphics[width=\linewidth]{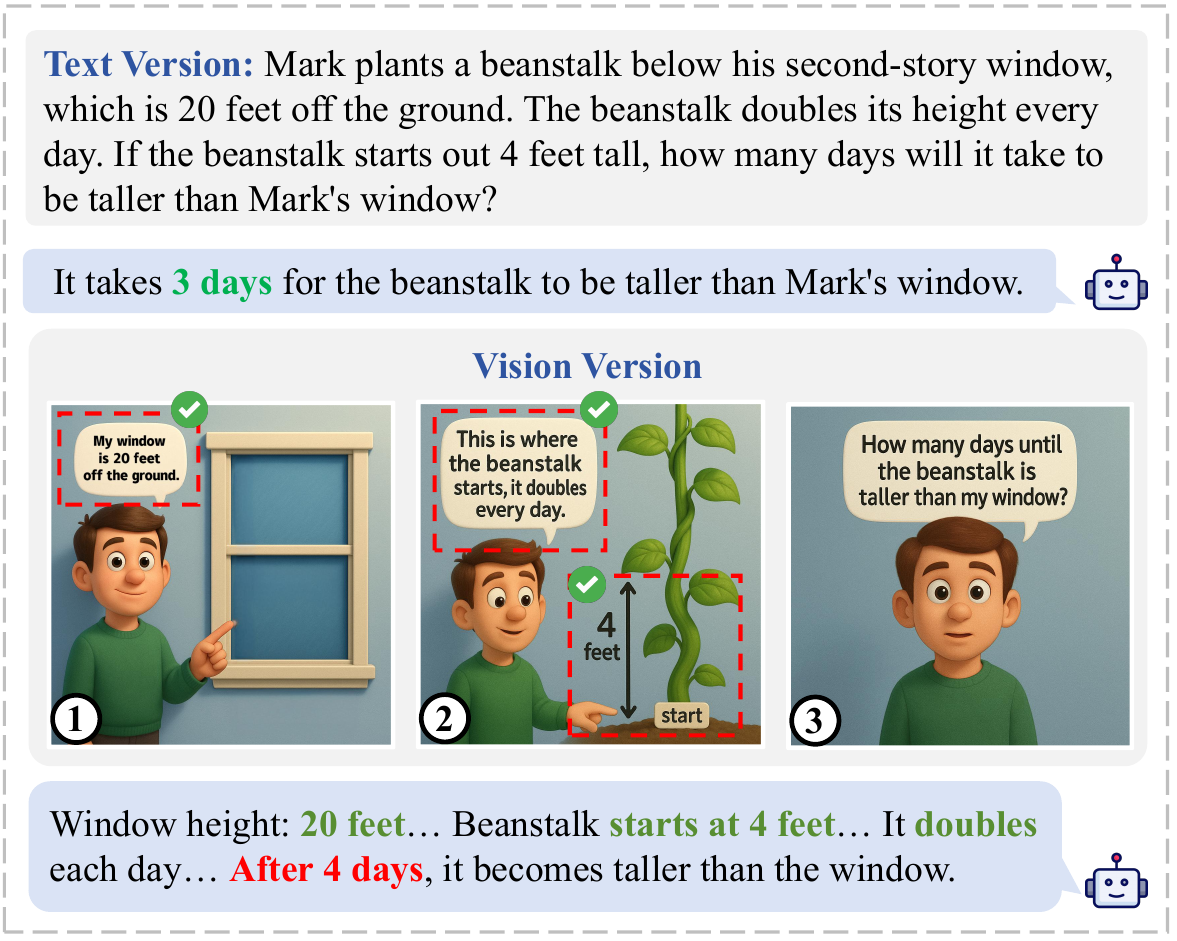}
  \caption{Illustration of the \textit{Seeing but Not Thinking} phenomenon. See Appendix~\ref{app:case_study} for details.}
  \label{fig:intro}
\end{wrapfigure}

Consider a simple scenario illustrated in Figure~\ref{fig:intro} and Appendix~\ref{app:case_study}: when presented with a grade-school mathematics problem \cite{cobbe2021training} as an image \cite{yuan2025gsm8k}, Qwen3-VL-30B-A3B \cite{bai2025qwen3vltechnicalreport} accurately extracts all numerical values and textual content, yet produces an incorrect answer due to reasoning errors. When the identical problem is presented as pure text, the same model solves it correctly with ease. We term this phenomenon \textbf{\textit{Seeing but Not Thinking}}: the model perceives visual content accurately but fails to reason correctly, despite possessing the requisite capability on semantically equivalent text inputs. This raises a fundamental question: {\textit{what factor causes multimodal MoE models to fail at reasoning when visual inputs are correctly perceived?}}

To systematically quantify this phenomenon while minimizing interference from perceptual errors, we construct rigorously controlled experiments based on the MATH500 \cite{hendrycks2021measuring} dataset. We render all pure-text problems as high-resolution images (detailed in Appendix~\ref{app:dataset_construction}) to ensure visual inputs are clear and legible.
To pinpoint the source of these failures, we conduct error analysis on samples that succeed on text but fail on images. The results are striking: 68.2\% to 73.1\% of failures stem from reasoning errors, while only 26.9\% to 31.8\% are attributable to perception errors (Table~\ref{tab:ocr_verification}).
This confirms that visual inputs degrade reasoning performance even when the content is correctly perceived. This finding aligns with conclusions from recent benchmarks \cite{zhang2024cross,van2025same}.

A natural hypothesis is cross-modal semantic alignment failure: visual information, though correctly perceived, may fail to align with the textual semantic space at the representation level. Prior work has demonstrated that dense-architecture VLMs achieve cross-modal semantic sharing \cite{wu2024semantic,shukor2024implicit}, but whether MoE-based VLMs possess the same property remains unexplored. We design cross-modal concept intervention experiments that manipulate hidden state representations across modalities.
Our results reveal a clear inverted U-shaped pattern across layers: intervention success rates are low in early layers where visual features have not yet aligned, peak in middle layers where semantic sharing occurs, and decline in terminal layers where output distributions are already determined. This finding confirms that MoE architectures also exhibit cross-modal semantic sharing, indicating that semantic alignment failure alone cannot account for the observed reasoning degradation.

If semantic alignment is not the primary bottleneck, what other factors might contribute? We examine the routing mechanism, the core component distinguishing MoE from dense models. Through systematic analysis of expert activation patterns, we uncover two critical findings. First, experts exhibiting high activation on visual inputs concentrate in early and terminal layers, while domain-specific reasoning experts cluster in middle layers. Second, image inputs induce significant routing divergence from text inputs precisely in these middle layers. Crucially, greater routing divergence correlates with lower reasoning accuracy across our controlled conditions. These observations lead us to propose the \textbf{\textit{Routing Distraction}} hypothesis: when processing visual inputs, the routing mechanism fails to adequately activate task-relevant reasoning experts, instead directing computation toward less suitable experts. The visual modality does not impair intrinsic reasoning capabilities; rather, it causes suboptimal expert selection, preventing full utilization of domain-specific reasoning capacity.

To validate this hypothesis, we design a routing-guided intervention that enhances domain expert activation during inference. The core idea is straightforward: if routing distraction significantly contributes to reasoning failures, then explicitly increasing the activation weights of domain-relevant experts should recover reasoning performance. We evaluate on three MoE models across six benchmarks spanning semantically equivalent and natural visual scenarios. The soft routing guidance strategy yields consistent improvements. Even on tasks involving complex geometric figures and function graphs where visual information cannot be replaced by text, enhancing domain expert activation helps models better integrate perception with reasoning, with gains of up to 3.17\%.
Further analysis reveals that domain expert identification is robust to information completeness of text references: as long as the reference elicits target domain reasoning patterns, identified experts transfer effectively to visual tasks with different information structures. This suggests that expert identification locates computational units responsible for cognitive functions, rather than memorizing sample-specific solutions.

Our contributions are threefold. First, we systematically characterize the \textit{Seeing but Not Thinking} phenomenon and demonstrate its prevalence across multiple state-of-the-art multimodal MoE models. Second, we provide mechanistic insights into this phenomenon through the \textit{Routing Distraction} hypothesis, revealing the layer-wise separation between visual and domain experts and the routing divergence induced by visual inputs. Third, we propose and validate a routing-guided intervention method that effectively mitigates routing distraction, achieving consistent improvements across diverse benchmarks and model scales.

\begin{figure*}[t]
  \centering
  \includegraphics[width=\textwidth]{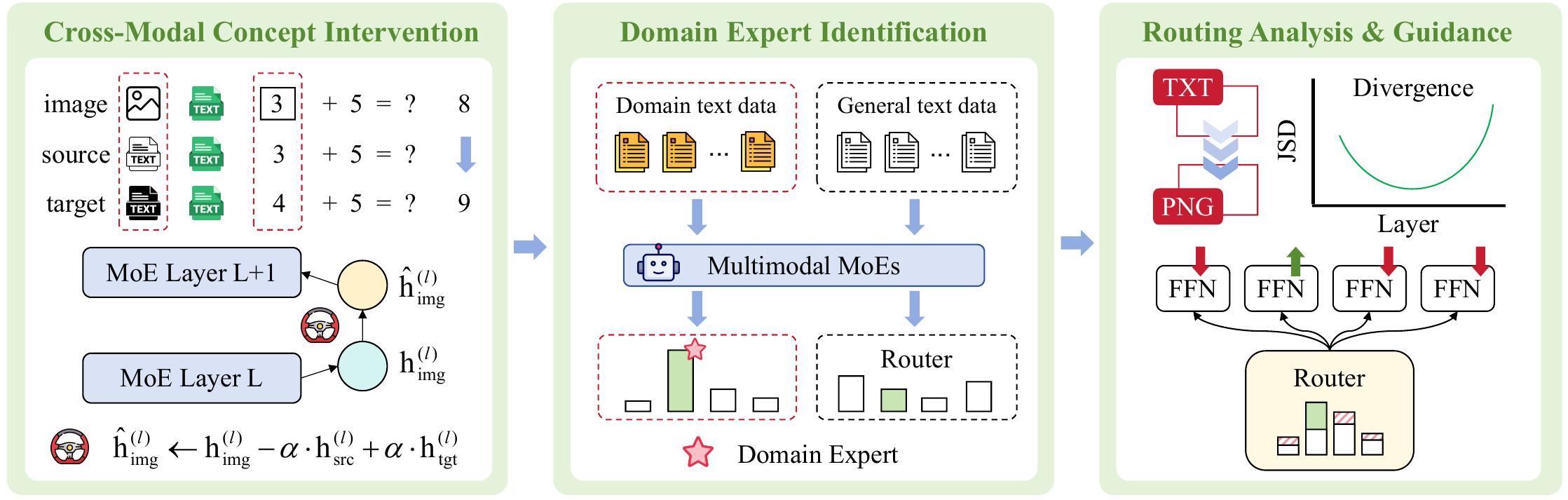}
  \caption{\textbf{Overview of our work.} We first conduct cross-modal concept intervention to verify semantic sharing in MoE architectures (left, \S\ref{sec:semantic_sharing}), then identify domain experts by comparing activation frequencies on domain-specific versus general data (middle, \S\ref{sec:expert_specialization}), and finally analyze routing divergence across modalities and apply routing guidance to enhance domain expert activation (right, \S\ref{sec:routing_divergence}-\S\ref{sec:method}).}
  \label{fig:overview}
\end{figure*}

\section{Related Work}
\paragraph{Multimodal Semantic Sharing} 
Vision-language models achieve cross-modal understanding by connecting visual encoders with large language models. Although image and text embeddings exhibit separated distributions in the shared space \cite{liang2022mind,schrodi2024two}, semantic sharing arise at deeper representation levels. Image representations can transfer to frozen language models through a single linear projection \cite{merullo2022linearly}, and representations from different modalities may converge toward a shared statistical model \cite{huh2024platonic}. In VLMs specifically, visual and text tokens activate similar LLM weights despite being distinct in representation space \cite{shukor2024implicit}, and semantically equivalent text-image inputs can be aligned into modality-invariant task vectors \cite{luo2024vision}. More broadly, shared semantic spaces have been observed across diverse modalities and languages \cite{wendler2024llamas,bandarkar2025multilingual,wu2024semantic}.

\paragraph{MoE Expert Specialization}
Mixture-of-Experts models scale effectively through sparse activation. Early MoE models such as OpenMoE \cite{xue2024openmoe} did not exhibit significant domain specialization due to the design of their load-balancing objectives. In modern MoE architectures, improved training strategies make domain specialization an expected outcome (see Appendix~\ref{app:theory}). This specialization manifests at multiple granularities: expert differentiation intensifies with layer depth \cite{lo2025closer}, cognitive experts control meta-level reasoning operations \cite{wang2025two}, and safety-related refusal behavior concentrates in a small subset of experts \cite{lai2025safex}.

\paragraph{Routing Intervention}
Inference-time routing intervention \cite{wu-etal-2025-sharp,ding2025d} has emerged as a promising direction. R2-T2 refines expert selection by shifting routing weights toward correctly predicted samples \cite{li2025r2}. SCMoE enhances reasoning by contrasting output distributions between selected and unselected experts \cite{shi2024unchosen}. SteerMoE identifies key experts through routing differences and adjusts routing logits for lightweight behavior control \cite{fayyaz2025steering}. Dynamic routing mechanisms can also enable complex tasks to activate more experts \cite{huang2024harder}.

\section{Analyzing Routing Distraction}

We systematically probe the mechanisms underlying the \textit{Seeing but Not Thinking} phenomenon. Our analysis proceeds in three stages: we first verify that cross-modal semantic sharing exists in MoE architectures (\S\ref{sec:semantic_sharing}), then examine the spatial distribution of specialized experts across layers (\S\ref{sec:expert_specialization}), and finally characterize routing behavior differences across modalities (\S\ref{sec:routing_divergence}). All experiments are conducted on Qwen3-VL-30B-A3B using controlled datasets that minimize perceptual confounds.

\subsection{Cross-Modal Semantic Sharing in MoE}
\label{sec:semantic_sharing}

\paragraph{Experimental Design.} To examine whether cross-modal semantic sharing exists in MoE architectures, we design a concept intervention experiment that directly alters hidden states across modalities. We construct an arithmetic completion task where the input consists of a digit image followed by a textual arithmetic expression (e.g., an image of "3" followed by "+ 2 ="). We extract hidden state vectors for the source digit $S$ and target digit $T$ from pure text inputs, denoted as $\mathbf{h}_{src}^{(l)}$ and $\mathbf{h}_{tgt}^{(l)}$ at layer $l$. We then perform the following intervention on the hidden states of image tokens:
\begin{equation}
\hat{\mathbf{h}}_{img}^{(l)} \leftarrow \mathbf{h}_{img}^{(l)} - \alpha \cdot \mathbf{h}_{src}^{(l)} + \alpha \cdot \mathbf{h}_{tgt}^{(l)}
\end{equation}
where $\alpha = 1$ controls intervention strength. This operation removes the source concept's semantic vector from the image representation while adding the target concept's vector. If the model's output changes to match the answer for the target digit, the intervention is deemed successful. We randomly generate 100 test instances with simple digit images to ensure perception is not a confounding factor.

\begin{figure*}[t]
  \centering
  \begin{subfigure}[b]{0.32\textwidth}
    \centering
    \includegraphics[width=\textwidth]{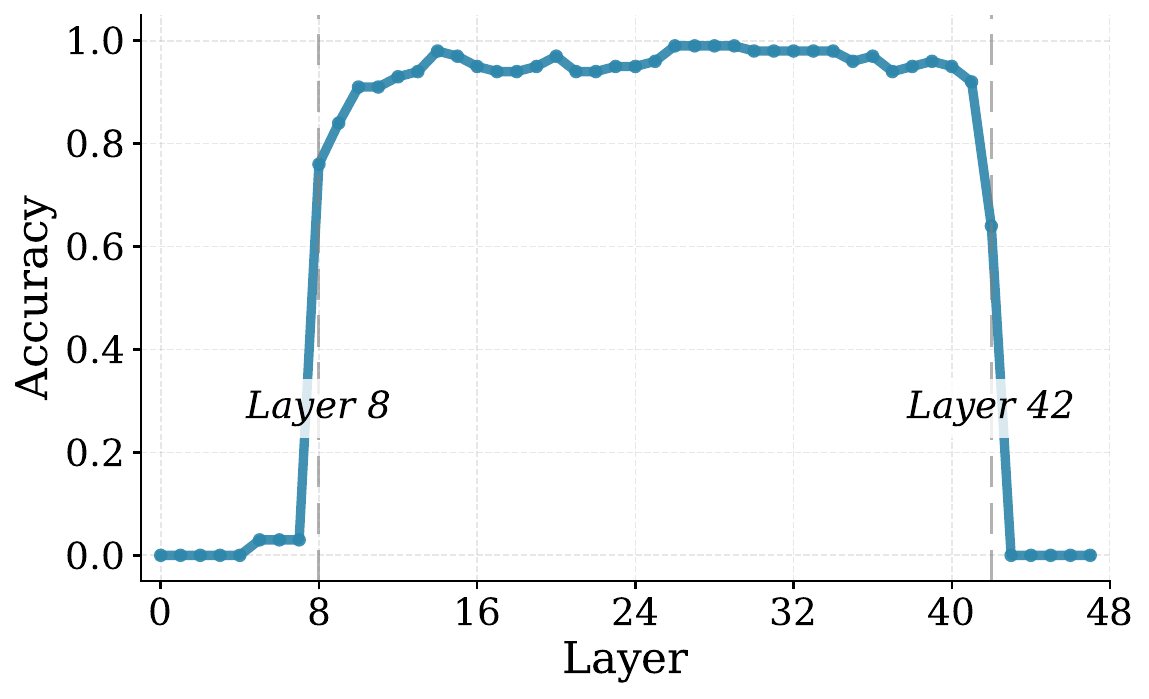}
    \caption{}
    \label{fig:intervention_sub}
  \end{subfigure}
  \hfill
  \begin{subfigure}[b]{0.32\textwidth}
    \centering
    \includegraphics[width=\textwidth]{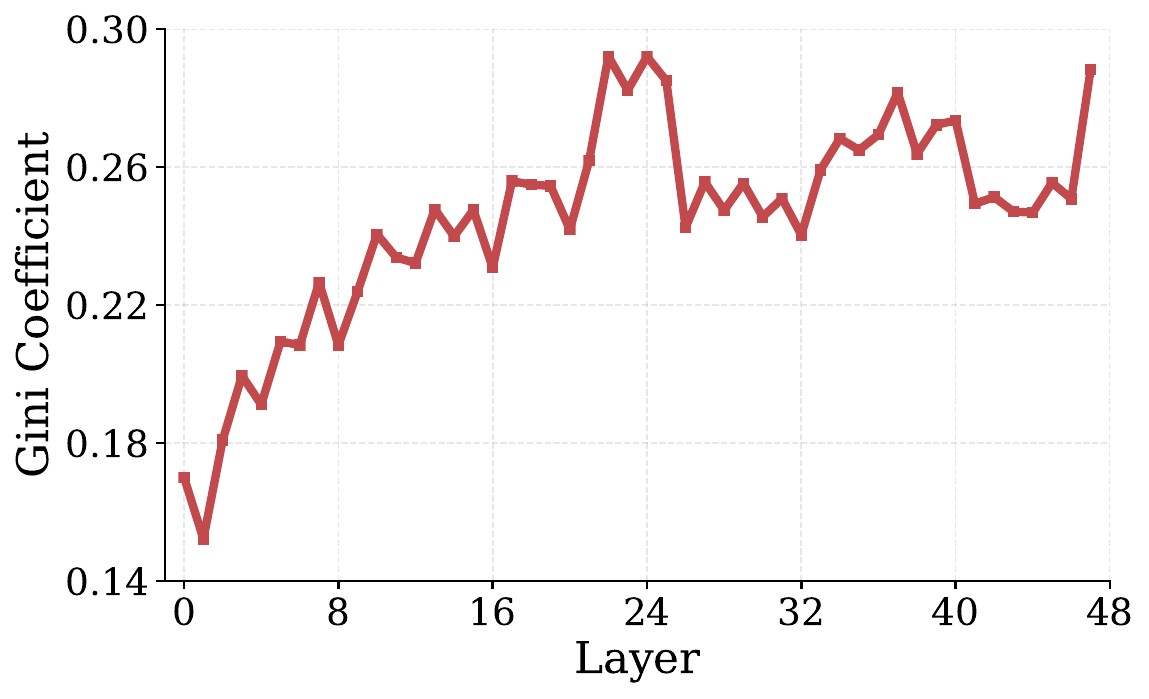}
    \caption{}
    \label{fig:gini_sub}
  \end{subfigure}
  \hfill
  \begin{subfigure}[b]{0.32\textwidth}
    \centering
    \includegraphics[width=\textwidth]{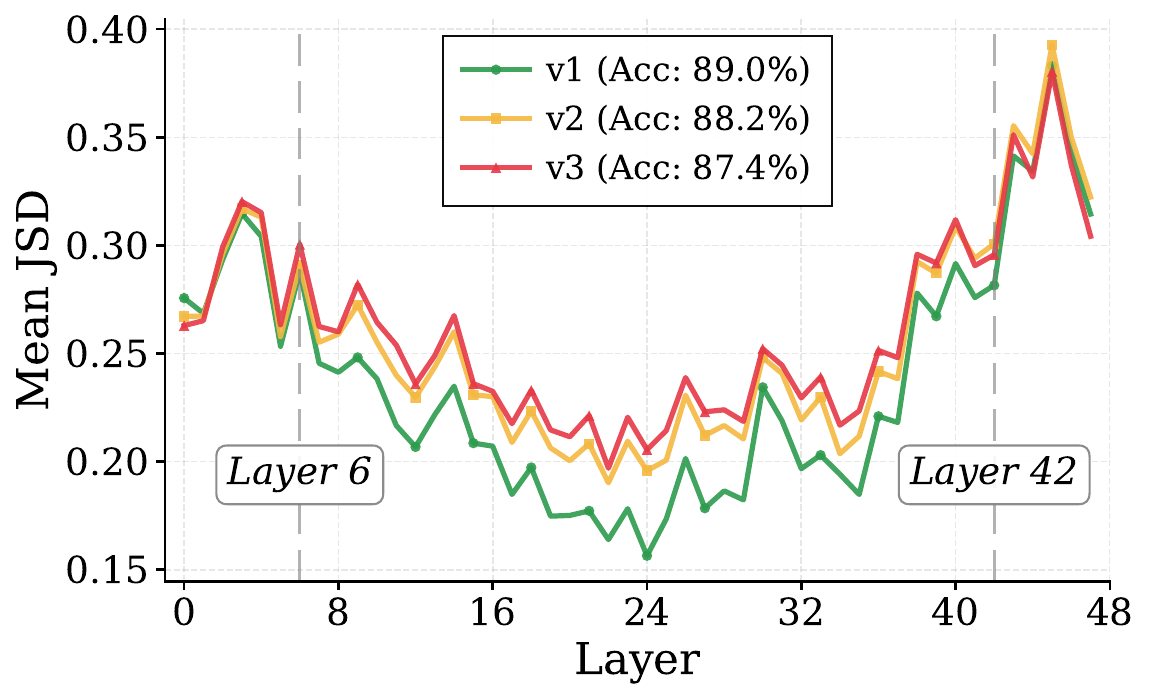}
    \caption{}
    \label{fig:jsd_sub}
  \end{subfigure}
  \caption{Analysis of routing mechanisms in multimodal MoE models. (a) Cross-modal semantic sharing verification showing inverted U-shaped intervention success rates. (b) Expert specialization quantification using Gini coefficients. (c) Routing divergence across modalities for three image versions.}
  \label{fig:intervention}
\end{figure*}

\paragraph{Results.} Figure~\ref{fig:intervention}(a) shows intervention success rates across layers. The results exhibit a clear inverted U-shaped pattern: early layers show low success rates, indicating that visual features have not yet aligned with the textual semantic space; middle layers (8–42) show significantly elevated success rates exceeding 90\%, suggesting that the two modalities achieve substantial semantic sharing in this region; terminal layers show a sharp decline in intervention effectiveness, likely because the model has already committed to its output distribution. This pattern confirms that MoE-based vision-language models exhibit cross-modal semantic sharing in middle layers, consistent with findings in dense architectures \cite{wu2024semantic,shukor2024implicit}. This result rules out semantic alignment failure as the sole explanation for \textit{Seeing but Not Thinking}, motivating us to explore other factors specific to MoE architectures.

\subsection{Layer-wise Expert Specialization}
\label{sec:expert_specialization}

Having established that semantic alignment is preserved across modalities, we now investigate another distinctive aspect of MoE architectures: expert specialization. We examine how different types of experts are distributed across layers and whether this distribution reveals structural patterns relevant to the observed phenomenon.

\paragraph{Quantifying Expert Specialization.} We first measure the degree of expert specialization at each layer using the Gini coefficient. For a sequence of length $L$, let $\mathbf{p}_{l,t}$ denote the expert probability vector for the $t$-th token at layer $l$. The average expert importance at layer $l$ is $\mathbf{q}_l = \frac{1}{L}\sum_{t=1}^{L}\mathbf{p}_{l,t}$, with $q_{l,i}$ representing the importance of expert $i$. The Gini coefficient is computed as:
\begin{equation}
\small
G_l = \frac{\sum_{i=1}^{E}\sum_{j=1}^{E}|q_{l,i} - q_{l,j}|}{2E\sum_{k=1}^{E}q_{l,k}}
\end{equation}
where $E$ is the total number of experts. Higher values indicate greater concentration of computation among fewer experts. As shown in Figure~\ref{fig:intervention}(b), early layers exhibit lower Gini coefficients while middle and terminal layers show elevated values, indicating that expert functional specialization intensifies in deeper layers.

\begin{figure*}[t]
  \centering
  \includegraphics[width=0.49\textwidth]{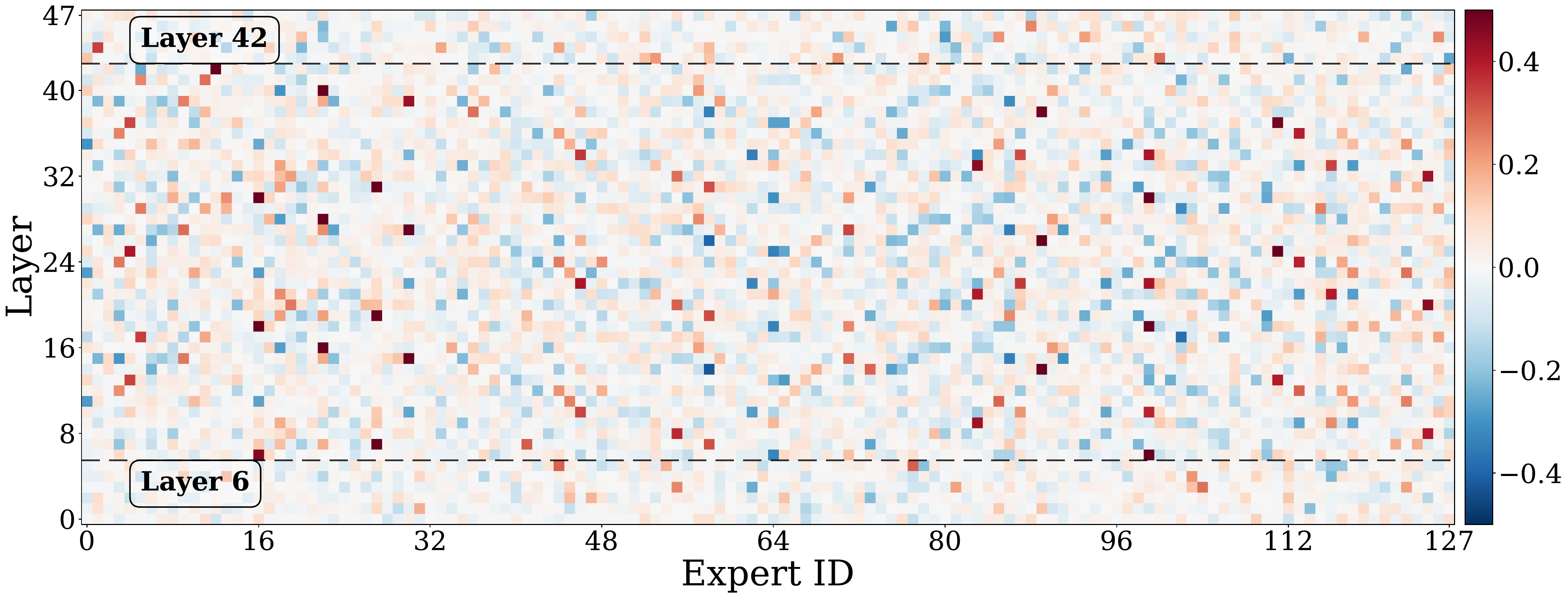}
  \hfill
  \includegraphics[width=0.49\textwidth]{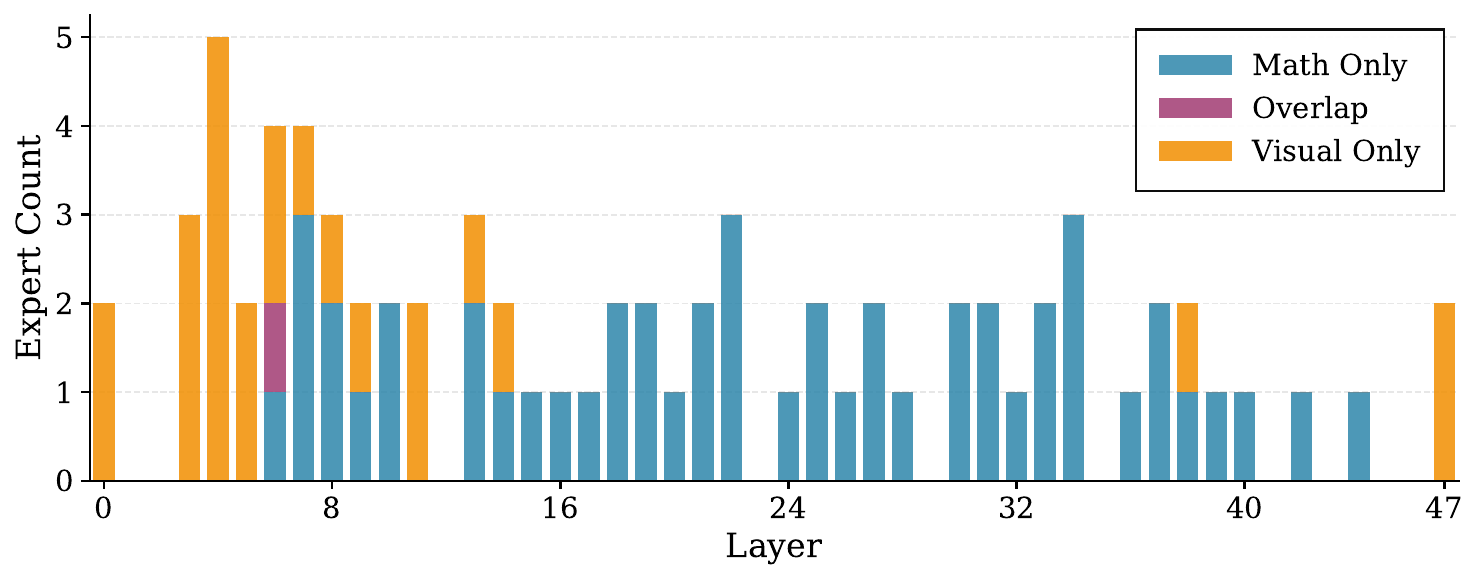}
  \caption{Layer-wise distribution of domain experts and visual experts. Left: Heatmap showing activation frequency differences (red: higher on math data; blue: higher on general data), with deep red concentrated in layers 6–42. Right: Expert counts per layer, where Overlap indicates experts identified as both math and visual experts.}
  \label{fig:expert_distribution}
\end{figure*}

\paragraph{Identifying Domain Experts.} To locate experts responsible for domain-specific reasoning, we compare activation frequencies between domain data and general data. Define the Top-K activation frequency of expert $E_{l,i}$ on dataset $D$ as:
\begin{equation}
\small
\Phi(E_{l,i}, D) = \frac{1}{N_D}\sum_{t \in D}\mathbf{1}[E_{l,i} \in \text{TopK}(x_t)]
\end{equation}
where $N_D$ is the total token count. The frequency difference $\Delta\Phi_{l,i} = \Phi(E_{l,i}, D_{\text{dom}}) - \Phi(E_{l,i}, D_{\text{gen}})$ captures domain-specific activation patterns. Experts with $\Delta\Phi_{l,i} > \tau$ are set as domain experts. Using GSM8K as domain data and Alpaca as general data with $\tau = 0.3$, we find that math experts cluster predominantly in middle layers (Figure~\ref{fig:expert_distribution}).

\paragraph{Identifying Visual Experts.} 

To locate experts associated with visual processing, we render Alpaca text as images and compute frequency differences between the image and text versions. With $\tau = 0.2$ (we also tried $\tau = 0.3$ but found too few experts), we identify visual experts that are preferentially activated for image inputs. These experts concentrate in early and terminal layers, exhibiting minimal overlap with math experts in middle layers.

\paragraph{Key Finding.} This analysis reveals a critical structural trait: visual experts and domain experts exhibit layer-wise separation. Visual experts cluster in early layers (for initial visual encoding) and terminal layers (preparing modality-specific outputs), while domain experts concentrate in middle layers where cross-modal semantic sharing occurs (\S\ref{sec:semantic_sharing}). This spatial segregation raises a natural question: when processing visual inputs, does the routing mechanism in middle layers adequately activate the domain experts necessary for reasoning? We investigate this question in the following section.

\subsection{Routing Divergence Across Modalities}
\label{sec:routing_divergence}

The layer separation between visual and domain experts implies routing behavior in middle layers may be critical for reasoning performance. We now directly examine how expert activation patterns differ between image and text inputs, and whether such differences correlate with reasoning degradation.

\paragraph{Experimental Setup.} We conduct analysis on the MATH500 dataset with semantically equivalent text and image versions. Three image versions (v1/v2/v3) are constructed with increasing visual complexity. Error analysis confirms that reasoning errors rather than perception errors dominate these failures (Table \ref{tab:ocr_verification}), ensuring strict control over perceptual factors. To explicitly measure how visual inputs alter expert selection compared to text inputs for the same problem, we calculate the divergence at the sample level. Let $\Phi_l(x)$ denote the expert activation frequency distribution at layer $l$ for input $x$. We quantify the average routing divergence as:
\begin{equation}
    \small
    Div_l = \frac{1}{N} \sum_{i=1}^{N} JSD(\Phi_l(x_i^{txt}), \Phi_l(x_i^{img}))
\end{equation}
where $N$ is the total sample count, and $JSD(\cdot)$ represents the Jensen-Shannon Divergence. JSD is computed over prompt-phase tokens only, as generation-phase tokens are exclusively textual.

\paragraph{Results.} Figure~\ref{fig:intervention}(c) shows JSD across layers for the three image versions. Two patterns emerge. First, JSD exhibits a U-shaped distribution: early and terminal layers show larger divergence (expected due to visual encoding and output preparation), while middle layers show smaller divergence. Second, and more critically, the three curves diverge primarily in the middle layer (6–42) region while remaining nearly identical in early and terminal layers. This indicates that visual complexity predominantly affects routing behavior in middle layers, precisely where domain experts concentrate.

\paragraph{Correlation with Reasoning Accuracy.} Despite similar perception error rates, the three image versions exhibit different reasoning performance: v1/v2/v3 achieve 89.0\%/88.2\%/87.4\% respectively, versus 92.8\% for pure text. Notably, versions with lower reasoning accuracy exhibit greater JSD in middle layers. This correlation suggests that routing divergence in middle layers, rather than perceptual quality, contributes to reasoning degradation.
\paragraph{The Routing Distraction Hypothesis.} Synthesizing the findings from \S\ref{sec:semantic_sharing}--\S\ref{sec:routing_divergence}, we propose the Routing Distraction hypothesis: when processing 
visual inputs, the MoE routing mechanism fails to adequately activate task-relevant domain experts in middle layers, instead directing computation toward other experts less suited for the reasoning task. 
This hypothesis explains why models can perceive correctly yet reason incorrectly: semantic alignment is preserved, but the computational resources required for reasoning are not fully mobilized. While our controlled analysis focuses on \textbf{\textit{semantically equivalent scenarios}} where perception is not a confounding factor, we expect this hypothesis to 
generalize to \textbf{\textit{natural visual scenarios}} involving complex figures and diagrams. In the following sections, we validate this hypothesis across both scenario types.

\section{Routing-Guided Intervention}
\label{sec:method}

Based on the Routing Distraction hypothesis, we propose a simple intervention strategy: if insufficient activation of domain experts contributes to reasoning failures, then explicitly enhancing their routing weights should recover performance.

\subsection{Domain Expert Identification}
\label{sec:expert_identification}

Following the method in \S\ref{sec:expert_specialization}, we find domain experts by comparing activation rates between domain-specific data and general-purpose data (Alpaca). This requires constructing text references that elicit the target domain's reasoning patterns.

Text reference construction depends on the scenario type. For semantically equivalent scenarios, we directly use the original text problems before rendering. For natural visual scenarios, we adopt task-appropriate proxies that elicit similar domain reasoning patterns, such as text-only problem versions or model-generated descriptions (\S\ref{sec:setup}).

Given text references, we compute $\Delta\Phi_{l,i}$ for each expert and apply threshold $\tau$ to obtain the domain expert set $\mathcal{E}_{\text{domain}}$. This procedure requires only 20 randomly sampled examples. Notably, the model need not solve these samples, nor must the text reference be strictly equivalent to the visual task. We analyze this robustness property in \S\ref{sec:robustness}.

\subsection{Routing Weight Adjustment}
\label{sec:routing_adjustment}

During inference, we enhance routing weights of identified domain experts. We investigate two strategies with a random baseline as control.

\paragraph{Soft Intervention.} We apply moderate additive enhancement to domain expert logits:
\begin{equation}
r'_{l,k} \leftarrow r_{l,k} + \lambda \cdot s(r_l), \quad \forall E_{l,k} \in \mathcal{E}_{\text{domain}}
\end{equation}
where $r_{l,k}$ is the original routing logit, $s(r_l)$ is the standard deviation of all expert logits at layer $l$, and $\lambda$ is the enhancement coefficient. This formulation preserves the router's flexibility to adjust based on specific inputs while systematically increasing domain expert activation probability.

\paragraph{Hard Intervention.} We force domain expert logits to the layer maximum:
\begin{equation}
r'_{l,k} \leftarrow \max_j r_{l,j} + \delta, \quad \forall E_{l,k} \in \mathcal{E}_{\text{domain}}
\end{equation}
where $\delta \sim \mathcal{N}(0, 10^{-4})$ introduces small perturbations to prevent identical logits, particularly relevant for architectures with Top-1 routing.

\paragraph{Random Baseline.} To verify that improvements stem from activating specific domain experts rather than routing perturbation itself, we randomly select the same number of experts at each layer and apply identical enhancement as Soft intervention.

\section{Experiments}
\label{sec:experiments}

\subsection{Experimental Setup}
\label{sec:setup}

\paragraph{Models.} We evaluate three multimodal MoE models spanning different scales: Qwen3-VL-30B-A3B \cite{bai2025qwen3vltechnicalreport}, Kimi-VL-16B-A3B \cite{team2025kimi}, and Llama4-Scout-109B-A17B \cite{llama4_2025}. Detailed model architectures are provided in Appendix~\ref{app:model_info}.

\paragraph{Benchmarks.} We construct two complementary evaluation scenarios. \textit{Semantically Equivalent Scenarios} include MATH500 and GPQA-Diamond (chemistry, physics subsets) \cite{rein2024gpqa} rendered as high-resolution images, providing controlled environments where perception is not a confounding factor. \textit{Natural Visual Scenarios} include MathVerse (vision-only version) \cite{zhang2024mathverse}, MATH-Vision \cite{wang2024measuring}, and GSM8K-V \cite{yuan2025gsm8k}, with text references constructed from the official text-only version for MathVerse, model-generated problem descriptions for MATH-Vision (Appendix~\ref{app:mathvision_desc}), and corresponding GSM8K text problems for GSM8K-V.

\begin{wraptable}{r}{0.53\linewidth}
\caption{Intervention configurations for each model. Intervention layers are specified as inclusive ranges.}
\label{tab:intervention_config}
\centering
\small
\begin{tabular}{cccc}
\toprule
\textbf{Model} & \makecell{\textbf{Intervention}\\\textbf{Layers}} & $\bm{\tau}$ & $\bm{\lambda}$ \\
\midrule
Kimi-VL-16B-A3B & [0, 20] & 0.3 & 0.5 \\
Qwen3-VL-30B-A3B & [6, 42] & 0.3 & 0.5 \\
Llama4-Scout-109B-A17B & [8, 40] & 0.3 & 0.2 \\
\bottomrule
\end{tabular}
\end{wraptable}

\paragraph{Implementation.} We use vLLM \cite{kwon2023efficient} with EasySteer \cite{xu2025easysteer} for inference-time intervention. Intervention configurations are summarized in Table~\ref{tab:intervention_config}. All models use $\tau = 0.3$ for domain expert identification. Intervention layers are selected based on findings from \S\ref{sec:expert_specialization}: middle layers for Qwen3-VL and Llama4-Scout, early and middle layers for Kimi-VL where domain experts emerge earlier. Due to slight non-determinism in vLLM, we run 16 trials with greedy decoding and report average accuracy verified by xVerify \cite{chen2025xverify}. Additional details are provided in Appendix~\ref{app:eval_details}.
\suppressfloats[t]

\subsection{Main Results}
\label{sec:main_results}

\begin{table*}[t]
    \caption{Accuracy (\%) across three multimodal MoE models and six benchmarks. Math, Chemistry, and Physics refer to image-rendered versions of MATH500 and GPQA-Diamond (chemistry and physics subsets) respectively. MATH-V denotes MATH-Vision.}
    \label{tab:main_results}
    \centering
    \small
    \newcommand{\datacolwidth}{1.5cm}  %
    \resizebox{\textwidth}{!}{%
    \begin{tabular}{cccccccc}
    \toprule
                             & \multicolumn{3}{c}{\textbf{Semantically Equivalent Scenarios}}                & \multicolumn{3}{c}{\textbf{Natural Visual Scenarios}}   &                           \\
                             \cmidrule(lr){2-4} \cmidrule(lr){5-7}
    \multirow{-2}{*}{\textbf{Method}} & Math    & Chemistry & Physics  & MathVerse & MATH-V & GSM8K-V & \multirow{-2}{*}{\textbf{Average}} \\
    \midrule
    \rowcolor{gray!13}
    \multicolumn{8}{l}{\textit{Kimi-VL-16B-A3B-Instruct}}                                                                                                                                                      \\
    \midrule
    Baseline                 & \makebox[\datacolwidth][c]{52.30} & \makebox[\datacolwidth][c]{25.54} & \makebox[\datacolwidth][c]{29.51} & \makebox[\datacolwidth][c]{35.41} & \makebox[\datacolwidth][c]{21.05} & \makebox[\datacolwidth][c]{8.11} & \makebox[\datacolwidth][c]{28.65} \\
    Random                   & \makebox[\datacolwidth][c]{51.64\makebox[0pt][l]{\textsuperscript{\textcolor{red}{-0.66}}}} & \makebox[\datacolwidth][c]{25.67\makebox[0pt][l]{\textsuperscript{\textcolor{lightgreen}{+0.13}}}} & \makebox[\datacolwidth][c]{28.78\makebox[0pt][l]{\textsuperscript{\textcolor{red}{-0.73}}}} & \makebox[\datacolwidth][c]{34.39\makebox[0pt][l]{\textsuperscript{\textcolor{red}{-1.02}}}} & \makebox[\datacolwidth][c]{19.74\makebox[0pt][l]{\textsuperscript{\textcolor{red}{-1.31}}}} & \makebox[\datacolwidth][c]{6.82\makebox[0pt][l]{\textsuperscript{\textcolor{red}{-1.29}}}} & \makebox[\datacolwidth][c]{27.84\makebox[0pt][l]{\textsuperscript{\textcolor{red}{-0.81}}}} \\
    Hard                     & \makebox[\datacolwidth][c]{53.06\makebox[0pt][l]{\textsuperscript{\textcolor{lightgreen}{+0.76}}}} & \makebox[\datacolwidth][c]{25.81\makebox[0pt][l]{\textsuperscript{\textcolor{lightgreen}{+0.27}}}} & \makebox[\datacolwidth][c]{\textbf{32.78}\makebox[0pt][l]{\textsuperscript{\textcolor{lightgreen}{+3.27}}}} & \makebox[\datacolwidth][c]{35.79\makebox[0pt][l]{\textsuperscript{\textcolor{lightgreen}{+0.38}}}} & \makebox[\datacolwidth][c]{21.71\makebox[0pt][l]{\textsuperscript{\textcolor{lightgreen}{+0.66}}}} & \makebox[\datacolwidth][c]{7.13\makebox[0pt][l]{\textsuperscript{\textcolor{red}{-0.98}}}} & \makebox[\datacolwidth][c]{29.38\makebox[0pt][l]{\textsuperscript{\textcolor{lightgreen}{+0.73}}}} \\
    Soft                     & \makebox[\datacolwidth][c]{\textbf{54.54}\makebox[0pt][l]{\textsuperscript{\textcolor{lightgreen}{+2.24}}}} & \makebox[\datacolwidth][c]{\textbf{27.89}\makebox[0pt][l]{\textsuperscript{\textcolor{lightgreen}{+2.35}}}} & \makebox[\datacolwidth][c]{32.49\makebox[0pt][l]{\textsuperscript{\textcolor{lightgreen}{+2.98}}}} & \makebox[\datacolwidth][c]{\textbf{38.58}\makebox[0pt][l]{\textsuperscript{\textcolor{lightgreen}{+3.17}}}} & \makebox[\datacolwidth][c]{\textbf{23.36}\makebox[0pt][l]{\textsuperscript{\textcolor{lightgreen}{+2.31}}}} & \makebox[\datacolwidth][c]{\textbf{9.17}\makebox[0pt][l]{\textsuperscript{\textcolor{lightgreen}{+1.06}}}} & \makebox[\datacolwidth][c]{\textbf{31.01}\makebox[0pt][l]{\textsuperscript{\textcolor{lightgreen}{+2.36}}}} \\
    \midrule
    \rowcolor{gray!13}
    \multicolumn{8}{l}{\textit{Qwen3-VL-30B-A3B-Instruct}}                                                                                                                                                     \\
    \midrule
    Baseline                 & \makebox[\datacolwidth][c]{88.20} & \makebox[\datacolwidth][c]{41.94} & \makebox[\datacolwidth][c]{75.58} & \makebox[\datacolwidth][c]{69.29} & \makebox[\datacolwidth][c]{55.92} & \makebox[\datacolwidth][c]{24.49} & \makebox[\datacolwidth][c]{59.24} \\
    Random                   & \makebox[\datacolwidth][c]{85.18\makebox[0pt][l]{\textsuperscript{\textcolor{red}{-3.02}}}} & \makebox[\datacolwidth][c]{36.56\makebox[0pt][l]{\textsuperscript{\textcolor{red}{-5.38}}}} & \makebox[\datacolwidth][c]{70.93\makebox[0pt][l]{\textsuperscript{\textcolor{red}{-4.65}}}} & \makebox[\datacolwidth][c]{67.78\makebox[0pt][l]{\textsuperscript{\textcolor{red}{-1.51}}}} & \makebox[\datacolwidth][c]{53.62\makebox[0pt][l]{\textsuperscript{\textcolor{red}{-2.30}}}} & \makebox[\datacolwidth][c]{23.42\makebox[0pt][l]{\textsuperscript{\textcolor{red}{-1.07}}}} & \makebox[\datacolwidth][c]{56.25\makebox[0pt][l]{\textsuperscript{\textcolor{red}{-2.99}}}} \\
    Hard                     & \makebox[\datacolwidth][c]{87.92\makebox[0pt][l]{\textsuperscript{\textcolor{red}{-0.28}}}} & \makebox[\datacolwidth][c]{33.33\makebox[0pt][l]{\textsuperscript{\textcolor{red}{-8.61}}}} & \makebox[\datacolwidth][c]{70.93\makebox[0pt][l]{\textsuperscript{\textcolor{red}{-4.65}}}} & \makebox[\datacolwidth][c]{68.40\makebox[0pt][l]{\textsuperscript{\textcolor{red}{-0.89}}}} & \makebox[\datacolwidth][c]{54.28\makebox[0pt][l]{\textsuperscript{\textcolor{red}{-1.64}}}} & \makebox[\datacolwidth][c]{23.65\makebox[0pt][l]{\textsuperscript{\textcolor{red}{-0.84}}}} & \makebox[\datacolwidth][c]{56.42\makebox[0pt][l]{\textsuperscript{\textcolor{red}{-2.82}}}} \\
    Soft                     & \makebox[\datacolwidth][c]{\textbf{89.42}\makebox[0pt][l]{\textsuperscript{\textcolor{lightgreen}{+1.22}}}} & \makebox[\datacolwidth][c]{\textbf{44.09}\makebox[0pt][l]{\textsuperscript{\textcolor{lightgreen}{+2.15}}}} & \makebox[\datacolwidth][c]{\textbf{76.74}\makebox[0pt][l]{\textsuperscript{\textcolor{lightgreen}{+1.16}}}} & \makebox[\datacolwidth][c]{\textbf{71.20}\makebox[0pt][l]{\textsuperscript{\textcolor{lightgreen}{+1.91}}}} & \makebox[\datacolwidth][c]{\textbf{57.57}\makebox[0pt][l]{\textsuperscript{\textcolor{lightgreen}{+1.65}}}} & \makebox[\datacolwidth][c]{\textbf{25.32}\makebox[0pt][l]{\textsuperscript{\textcolor{lightgreen}{+0.83}}}} & \makebox[\datacolwidth][c]{\textbf{60.72}\makebox[0pt][l]{\textsuperscript{\textcolor{lightgreen}{+1.48}}}} \\
    \midrule
    \rowcolor{gray!13}
    \multicolumn{8}{l}{\textit{Llama4-Scout-109B-A17B-Instruct}}                                                                                                                                               \\
    \midrule
    Baseline                 & \makebox[\datacolwidth][c]{77.95} & \makebox[\datacolwidth][c]{42.20} & \makebox[\datacolwidth][c]{55.60} & \makebox[\datacolwidth][c]{56.09} & \makebox[\datacolwidth][c]{32.24} & \makebox[\datacolwidth][c]{23.50} & \makebox[\datacolwidth][c]{47.93} \\
    Random                   & \makebox[\datacolwidth][c]{77.80\makebox[0pt][l]{\textsuperscript{\textcolor{red}{-0.15}}}} & \makebox[\datacolwidth][c]{41.40\makebox[0pt][l]{\textsuperscript{\textcolor{red}{-0.80}}}} & \makebox[\datacolwidth][c]{54.65\makebox[0pt][l]{\textsuperscript{\textcolor{red}{-0.95}}}} & \makebox[\datacolwidth][c]{54.57\makebox[0pt][l]{\textsuperscript{\textcolor{red}{-1.52}}}} & \makebox[\datacolwidth][c]{31.58\makebox[0pt][l]{\textsuperscript{\textcolor{red}{-0.66}}}} & \makebox[\datacolwidth][c]{22.82\makebox[0pt][l]{\textsuperscript{\textcolor{red}{-0.68}}}} & \makebox[\datacolwidth][c]{47.14\makebox[0pt][l]{\textsuperscript{\textcolor{red}{-0.79}}}} \\
    Hard                     & \makebox[\datacolwidth][c]{77.62\makebox[0pt][l]{\textsuperscript{\textcolor{red}{-0.33}}}} & \makebox[\datacolwidth][c]{37.16\makebox[0pt][l]{\textsuperscript{\textcolor{red}{-5.04}}}} & \makebox[\datacolwidth][c]{46.73\makebox[0pt][l]{\textsuperscript{\textcolor{red}{-8.87}}}} & \makebox[\datacolwidth][c]{53.55\makebox[0pt][l]{\textsuperscript{\textcolor{red}{-2.54}}}} & \makebox[\datacolwidth][c]{30.59\makebox[0pt][l]{\textsuperscript{\textcolor{red}{-1.65}}}} & \makebox[\datacolwidth][c]{22.44\makebox[0pt][l]{\textsuperscript{\textcolor{red}{-1.06}}}} & \makebox[\datacolwidth][c]{44.68\makebox[0pt][l]{\textsuperscript{\textcolor{red}{-3.25}}}} \\
    Soft                     & \makebox[\datacolwidth][c]{\textbf{79.20}\makebox[0pt][l]{\textsuperscript{\textcolor{lightgreen}{+1.25}}}} & \makebox[\datacolwidth][c]{\textbf{43.01}\makebox[0pt][l]{\textsuperscript{\textcolor{lightgreen}{+0.81}}}} & \makebox[\datacolwidth][c]{\textbf{56.98}\makebox[0pt][l]{\textsuperscript{\textcolor{lightgreen}{+1.38}}}} & \makebox[\datacolwidth][c]{\textbf{57.11}\makebox[0pt][l]{\textsuperscript{\textcolor{lightgreen}{+1.02}}}} & \makebox[\datacolwidth][c]{\textbf{33.88}\makebox[0pt][l]{\textsuperscript{\textcolor{lightgreen}{+1.64}}}} & \makebox[\datacolwidth][c]{\textbf{24.41}\makebox[0pt][l]{\textsuperscript{\textcolor{lightgreen}{+0.91}}}} & \makebox[\datacolwidth][c]{\textbf{49.10}\makebox[0pt][l]{\textsuperscript{\textcolor{lightgreen}{+1.17}}}} \\
    \bottomrule
    \end{tabular}%
    }
    \end{table*}

Table~\ref{tab:main_results} presents reasoning accuracy across all models and benchmarks. We analyze the results from three perspectives.

\paragraph{Effectiveness in Semantically Equivalent Scenarios.} These setup directly test the Routing Distraction hypothesis by isolating reasoning from perception. The Soft intervention achieves consistent improvements across all models. Kimi-VL improves from 52.30\% to 54.54\% (+2.24\%) on math and from 29.51\% to 32.49\% (+2.98\%) on physics. Qwen3-VL shows gains of 1.22\% on math and 2.15\% on chemistry. These results confirm that enhancing domain expert activation effectively mitigates routing distraction when perception is fixed.

\paragraph{Generalization to Natural Visual Scenarios.} A key question is whether routing distraction extends to scenarios with complex visual elements. Kimi-VL rises from 35.41\% to 38.58\% (+3.17\%) on MathVerse; Qwen3-VL improves from 55.92\% to 57.57\% (+1.65\%) on MATH-Vision. These gains indicate  enhancing domain expert activation helps models better integrate visual information. Improvements on GSM8K-V are modest, suggesting that when task difficulty concentrates in the perception phase (extracting information from multiple scene images), routing guidance offers less benefit.

\paragraph{Comparison of Intervention Strategies.} Soft intervention performs most consistently, balancing enhanced domain expert activation with preserved routing flexibility. Hard intervention shows mixed results: while it sometimes yields improvements (e.g., Kimi-VL on physics), it causes severe degradation on Llama4-Scout, even underperforming the Random baseline. We attribute this to Llama4's Top-1 routing mechanism, where forcing expert logits throughout generation frequently disrupts the original routing decisions. The Random baseline generally produces no improvement, confirming that gains stem from activating correct domain experts rather than routing perturbation.

\subsection{Analysis}
\label{sec:analysis}
\paragraph{Robustness of Expert Identification.}
\label{sec:robustness}
A critical question is whether domain expert identification demands semantically equivalent text references. This is particularly relevant for natural visual benchmarks where perfect text equivalents do not exist. Taking MathVerse as an example, its official text-only version cannot fully convey spatial relationships and geometric configurations, rendering many problems unsolvable from text alone. Indeed, Qwen3-VL achieves 67.26\% accuracy on text-only, while achieving 69.29\% on vision-only, indicating that visual information provides independent value.

However, experts identified using this partial text reference still yield substantial improvements: vision-only accuracy rises to 71.20\% (+1.91\%) with Soft intervention. This result carries two implications. First, visual information provides unique value that text misses; routing guidance helps models leverage this information rather than regressing to text-based reasoning paths. Second, expert identification is robust to information completeness. The procedure locates computational units responsible for domain-specific cognitive functions, not sample-specific solution paths. As long as the text reference elicits target domain reasoning patterns, identified experts transfer effectively to visual tasks with different information structures.

\begin{table}[t]
\setlength{\tabcolsep}{4.5pt}%
\begin{minipage}[b]{0.472\linewidth}

\caption{Text-only and vision-only accuracy on MathVerse, with guidance using experts identified from MathVerse text-only vs. GSM8K. Superscripts indicate the change relative to the vision-only accuracy.}
\label{tab:domain_specificity}
\centering
\small
\resizebox{\linewidth}{!}{%
\begin{tabular}{ccccc}
\toprule
\multirow{2.5}{*}{\textbf{Model}} & \multirow{2.5}{*}{\textbf{Text}} & \multirow{2.5}{*}{\textbf{Vision}} & \multicolumn{2}{c}{\textbf{w/ Guidance}} \\
\cmidrule(lr){4-5}
& & & $\text{Ref}_{\text{MathVerse}}$ & $\text{Ref}_{\text{GSM8K}}$ \\
\midrule
Kimi-VL & 45.94 & 35.41 & \textbf{38.58}\textsuperscript{\textcolor{lightgreen}{+3.17}} & 37.31\textsuperscript{\textcolor{lightgreen}{+1.90}} \\
Qwen3-VL & 67.26 & 69.29 & \textbf{71.20}\textsuperscript{\textcolor{lightgreen}{+1.91}} & 69.54\textsuperscript{\textcolor{lightgreen}{+0.25}} \\
Llama4 & 63.32 & 56.09 & \textbf{57.11}\textsuperscript{\textcolor{lightgreen}{+1.02}} & 55.08\textsuperscript{\textcolor{red}{-1.01}} \\
\bottomrule
\end{tabular}%
}

\par\vspace{0pt}%
\end{minipage}\hfill
\begin{minipage}[b]{0.488\linewidth}

\caption{Impact of intervention layer range on average accuracy. Middle: intervening only on middle layers; w/ Early: adding early layers; w/ Late: adding late layers. For Kimi-VL, middle layers refer to layers 6-20.}
\label{tab:layer_ablation}
\centering
\small
\resizebox{\linewidth}{!}{%
\begin{tabular}{ccccc}
\toprule
\multirow{2.5}{*}{\textbf{Model}} & \multirow{2.5}{*}{\textbf{Baseline}} & \multirow{2.5}{*}{\textbf{w/ Early}} & \multirow{2.5}{*}{\textbf{Middle}} & \multirow{2.5}{*}{\textbf{w/ Late}} \\
\cmidrule[0pt](lr){2-5}
& & & & \\
\midrule
Kimi-VL & 28.65 & \textbf{31.01} & 29.82 & 29.35 \\
Qwen3-VL & 59.24 & 59.76 & \textbf{60.72} & 58.97 \\
Llama4 & 47.93 & 46.85 & \textbf{49.10} & 47.82 \\
\bottomrule
\end{tabular}%
}

\par\vspace{0pt}
\end{minipage}
\end{table}
\begin{wrapfigure}{r}{0.45\linewidth}
  \centering
  \includegraphics[width=\linewidth]{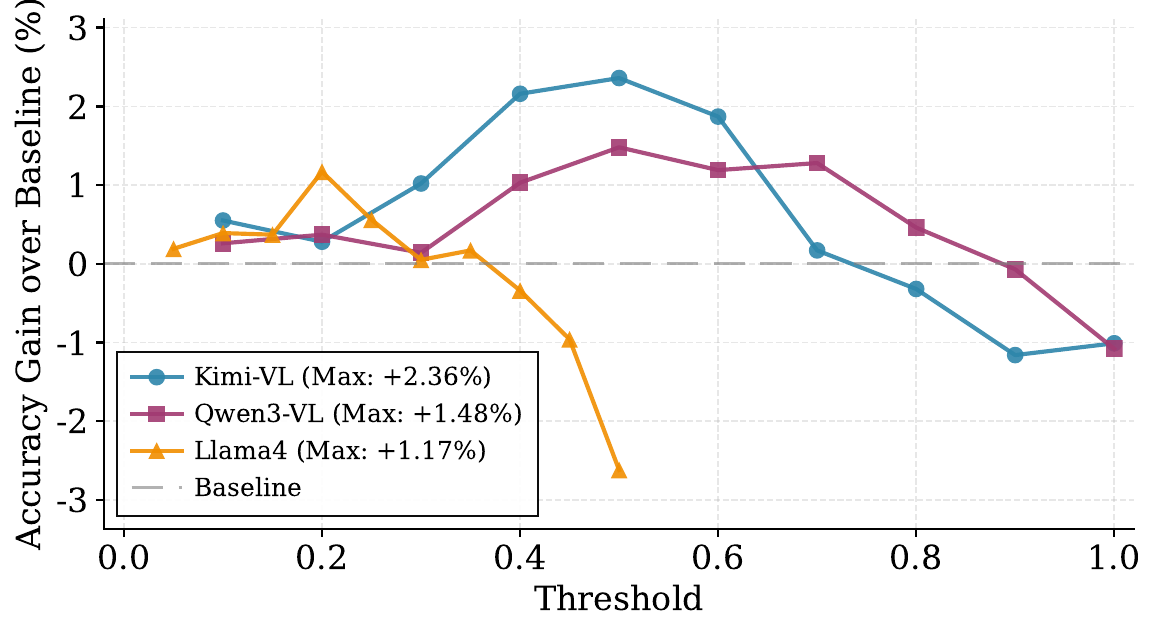}
  \caption{Effect of enhancement coefficient $\lambda$ on reasoning accuracy gains across three models. }
  \label{fig:threshold_accuracy}
\end{wrapfigure}
\paragraph{Domain Specificity.}
While expert identification is robust to information completeness, it remains sensitive to domain mismatch. When using GSM8K (elementary arithmetic) instead of MathVerse text-only as the reference for MathVerse evaluation, improvements diminish substantially or even reverse. As shown in Table \ref{tab:domain_specificity}, Qwen3-VL's vision-only accuracy rises from 69.29\% to 71.20\% (+1.91\%) with MathVerse-matched guidance, but only reaches 69.54\% (+0.25\%) with GSM8K-mismatched guidance. Notably, Llama4 shows degradation: from 56.09\% baseline to 57.11\% (+1.02\%) with matched guidance, but drops to 55.08\% (-1.01\%) when mismatched. Although both datasets involve mathematics, GSM8K emphasizes arithmetic operations while MathVerse requires geometric reasoning and function analysis. The experts activated by these distinct patterns do not fully overlap, confirming that effective intervention requires domain-matched references.

\paragraph{Layer Selection.}
We examine the impact of intervention layer range. As shown in Table~\ref{tab:layer_ablation}, Qwen3-VL and Llama4-Scout achieve optimal performance when intervening only on middle layers. Adding early layers degrades performance, likely because visual experts in these layers handle necessary visual feature extraction; premature intervention disrupts this processing. Kimi-VL exhibits a different pattern, benefiting from early and middle layer intervention. Analysis reveals that Kimi-VL's domain experts and cross-modal semantic sharing both emerge at earlier layers, making early intervention beneficial for this architecture.

\paragraph{Intervention Strength.}
Figure~\ref{fig:threshold_accuracy} shows the effect of enhancement coefficient $\lambda$ in Soft intervention. Kimi-VL and Qwen3-VL achieve optimal performance with $\lambda \in [0.4, 0.6]$; excessive values degrade accuracy by overriding input-specific routing decisions. Llama4 requires weaker intervention ($\lambda=0.2$), due to its Top-1 routing mechanism where activating only one expert per layer makes routing decisions more sensitive to logit changes.

\section{Conclusion}
\label{sec:conclusion}
This paper investigates the \textit{Seeing but Not Thinking} phenomenon in multimodal MoE models. We establish that cross-modal semantic sharing exists in MoE architectures, ruling out alignment failure as the sole explanation. Our analysis reveals layer-wise separation between visual and domain experts, with image inputs inducing routing divergence in middle layers that correlates with reasoning degradation. Based on these findings, we propose the Routing Distraction hypothesis and validate it through routing-guided intervention, achieving consistent improvements across three models and six benchmarks, with gains of up to 3.17\% on complex visual reasoning tasks. Our work provides both mechanistic insights into multimodal reasoning failures and a practical method for mitigation.

\section*{Limitations}
This work has several limitations. First, routing guidance cannot address perceptual errors; when visual information is incorrectly extracted, enhancing domain expert activation provides no benefit. The method targets the specific failure mode of correct perception coupled with reasoning failure, not the broader challenge of visual understanding.

Second, the current approach requires task-specific configuration. Domain expert identification relies on constructing appropriate text references, and optimal intervention layers and strengths must be determined empirically for each model-task combination. Developing adaptive methods that automatically identify relevant experts and calibrate intervention parameters would substantially improve practical applicability.

Third, our cross-modal semantic sharing experiments verify alignment for simple numerical concepts. Whether complex visual concepts (spatial relationships, geometric configurations, abstract diagrams) achieve equal alignment quality remains unclear. Insufficient alignment for complex concepts could constitute an additional factor contributing to reasoning failures beyond routing distraction.

Finally, while we demonstrate correlation between routing divergence and reasoning degradation, establishing strict causality would require more controlled interventions. The consistent improvements from routing guidance provide supporting evidence, but the precise causal mechanisms underlying expert selection in multimodal contexts merit further investigation.

\section*{Acknowledgments}
This work is supported by the National Natural Science Foundation of China (No. 62376245), the Key Research and Development Program of Zhejiang Province, China (No. 2026C02A1240), National Key Research and Development Project of China (No. 2018AAA0101900), and MOE Engineering Research Center of Digital Library.
\bibliographystyle{plainnat}
\bibliography{references}
\clearpage
\appendix
\section*{Supplementary Material}
\subsection*{Contents}
\noindent\hyperref[sec:discussion]{\ref*{sec:discussion}\quad \nameref*{sec:discussion}}\dotfill\pageref*{sec:discussion}\par
\noindent\hyperref[app:case_study]{\ref*{app:case_study}\quad \nameref*{app:case_study}}\dotfill\pageref*{app:case_study}\par
\noindent\hyperref[app:dataset_construction]{\ref*{app:dataset_construction}\quad \nameref*{app:dataset_construction}}\dotfill\pageref*{app:dataset_construction}\par
\noindent\hyperref[app:model_info]{\ref*{app:model_info}\quad \nameref*{app:model_info}}\dotfill\pageref*{app:model_info}\par
\noindent\hyperref[app:benchmarks]{\ref*{app:benchmarks}\quad \nameref*{app:benchmarks}}\dotfill\pageref*{app:benchmarks}\par
\noindent\hyperref[app:theory]{\ref*{app:theory}\quad \nameref*{app:theory}}\dotfill\pageref*{app:theory}\par
\noindent\hyperref[app:eval_details]{\ref*{app:eval_details}\quad \nameref*{app:eval_details}}\dotfill\pageref*{app:eval_details}\par
\medskip
\section{Discussion}
\label{sec:discussion}
\paragraph{Scope of Routing Distraction.} The routing distraction hypothesis explains cases where perception succeeds but reasoning fails. However, it does not address perceptual errors, nor does it claim to be the sole factor underlying multimodal reasoning failures. For tasks where difficulty concentrates in perception (e.g., GSM8K-V requiring information extraction from multiple scene images), routing guidance provides limited benefit. The phenomenon likely arises from multiple interacting factors, with routing distraction being one identifiable and addressable component.
\paragraph{Generality Across Architectures.} Our experiments span three models with different scales and routing mechanisms (Top-K vs. Top-1), demonstrating consistent improvements. However, the optimal intervention parameters (layers, strength) vary across architectures, reflecting differences in where domain experts emerge and how routing decisions propagate. Future work could investigate whether architectural properties predict optimal intervention configurations.
\paragraph{From Inference-Time Remedy to Training-Time Solution.} While routing-guided intervention effectively mitigates routing distraction, it remains a post-hoc remedy requiring domain expert identification and hyperparameter tuning for each task. A more fundamental solution lies in the training phase: designing architectures or training objectives that encourage routers to allocate experts based on task requirements rather than input modality. Potential directions include routing regularization that penalizes modality-dependent expert selection, auxiliary losses that align routing distributions across semantically equivalent multimodal inputs or better behavior internalization \cite{lu2026skill0,du2026s3}.

\section{Case Study}
\label{app:case_study}

Figure~\ref{fig:intro} presents a sample from the GSM8K dataset \cite{cobbe2021training}, with the corresponding image from GSM8K-V \cite{yuan2025gsm8k}. The model successfully extracts all necessary information and the question to be answered, but makes a reasoning error during inference by incorrectly counting the initial value as a separate day. Table~\ref{tab:case_study} shows the complete model responses before and after routing-guided intervention.

\begin{table*}[t]
\caption{Complete model responses for the problem shown in Figure~\ref{fig:intro}. Despite successfully extracting all necessary information, the baseline response incorrectly counts the initial value as a separate day (Day 1), leading to an off-by-one error. The response with routing-guided intervention correctly starts the count from Day 0 and produces the correct answer.}
\label{tab:case_study}
\centering
\small
\begin{tabular}{p{0.95\textwidth}}
\toprule
\textbf{Baseline Response (Incorrect: Answer is 4)} \\
\midrule
Of course, let's solve this problem step-by-step.

The scenario involves a beanstalk that \textcolor{lightgreen}{doubles in height every day} and \textcolor{lightgreen}{starts at 4 feet tall}. We need to find out how many days it will take for the beanstalk to grow taller than a window \textcolor{lightgreen}{located 20 feet off the ground}.

We can calculate the height of the beanstalk each day until its height exceeds 20 feet.

- \textbf{\textcolor{red}{Day 1:}} The beanstalk is $4$ feet tall.

- \textbf{\textcolor{red}{Day 2:}} It doubles from Day 1: $4 \times 2 = 8$ feet tall.

- \textbf{\textcolor{red}{Day 3:}} It doubles from Day 2: $8 \times 2 = 16$ feet tall.

- \textbf{\textcolor{red}{Day 4:}} It doubles from Day 3: $16 \times 2 = 32$ feet tall.

Now we compare these heights to the window:

- On \textcolor{red}{Day 3}, the beanstalk is 16 feet tall, which is less than 20 feet (the window).

- On \textcolor{red}{Day 4}, the beanstalk is 32 feet tall, which is greater than 20 feet.

Therefore, on the \textcolor{red}{fourth day}, the beanstalk becomes taller than the window.

It takes $\boxed{\textcolor{red}{4}}$ days for the beanstalk to be taller than the window. \\
\midrule
\textbf{With Routing Intervention (Correct: Answer is 3)} \\
\midrule
Of course, let's solve this problem step-by-step.

The scenario involves a beanstalk that doubles in height every day and starts at 4 feet tall. We need to find out how many days it will take for the beanstalk to grow taller than a window located 20 feet off the ground.

We can model the growth of the beanstalk using an exponential function:

- \textbf{Initial Height (Day 0):} $h_0 = 4$ feet.

- \textbf{Growth Rate:} The plant doubles its size each day, so we multiply by 2 daily.

- \textbf{Height on Day n:} $h_n = h_0 \times 2^n$

Let's calculate the height of the beanstalk over several consecutive days until it surpasses the 20-foot mark:

---

\textbf{Step 1: Calculate the height after each day}

- \textbf{After Day 1:} $h_1 = 4 \text{ ft} \times 2^1 = 8 \text{ ft}$

- \textbf{After Day 2:} $h_2 = 4 \text{ ft} \times 2^2 = 4 \text{ ft} \times 4 = 16 \text{ ft}$

- \textbf{After Day 3:} $h_3 = 4 \text{ ft} \times 2^3 = 4 \text{ ft} \times 8 = 32 \text{ ft}$

---

\textbf{Step 2: Compare with the window height}

Now, compare these heights to the window's height of 20 feet:

- After Day 1 ($8$ ft), the beanstalk is shorter than the window ($8 < 20$).

- After Day 2 ($16$ ft), the beanstalk is still shorter than the window ($16 < 20$).

- After Day 3 ($32$ ft), the beanstalk is now taller than the window ($32 > 20$).

Therefore, the first time the beanstalk exceeds the height of the window is on the third day.

It takes $\boxed{3}$ days for the beanstalk to be taller than the window. \\
\bottomrule
\end{tabular}
\end{table*}

\section{Construction and Validation of Semantically Equivalent Dataset}
\label{app:dataset_construction}

We constructed semantically equivalent images based on the MATH500 dataset. To analyze the relationship between routing divergence and reasoning accuracy in \S\ref{sec:routing_divergence}, we created three versions with different visual styles.

\paragraph{Rendering Method.} We used LaTeX to render and crop text problems into high-resolution images (dpi=200). For mathematical formulas, we used the default LaTeX font; for code segments, we used JetBrainsMono-Regular. We constructed three different versions:

\begin{itemize}
\item \textbf{v1}: Plain text uses the SimHei font (printed style)
\item \textbf{v2}: Plain text uses the Caveat font (handwritten style)  
\item \textbf{v3}: Adds a light gray grid background to v2
\end{itemize}

\begin{figure}[htb]
  \centering
  \includegraphics[width=0.6\linewidth]{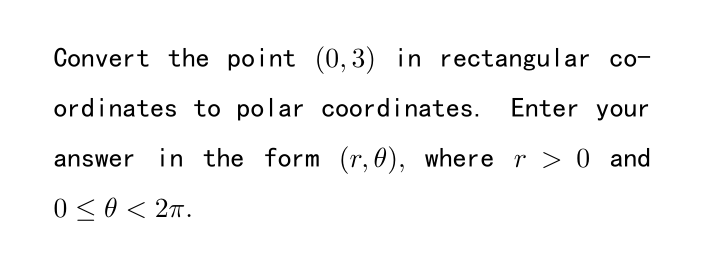}
  \includegraphics[width=0.6\linewidth]{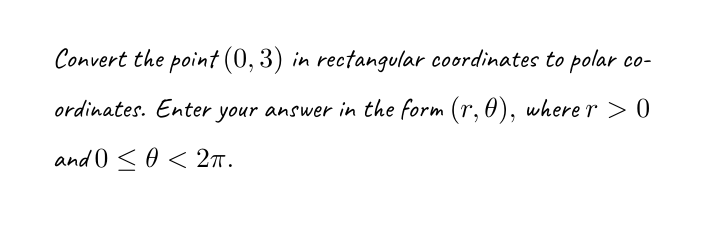}
  \includegraphics[width=0.6\linewidth]{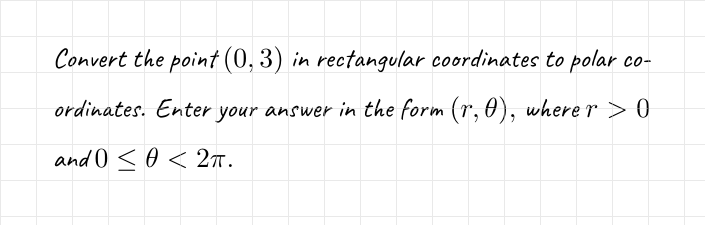}
  \caption{Examples of the three semantically equivalent image versions (v1, v2, v3 from top to bottom) for the same text problem.}
  \label{fig:dataset_samples}
\end{figure}

Images of the three versions corresponding to the same text are padded to maintain identical dimensions. Examples are shown in Figure~\ref{fig:dataset_samples}. The Alpaca dataset in \S\ref{sec:expert_specialization} was rendered following the v1 approach, and GPQA-Diamond in \S\ref{sec:experiments} was rendered following the v2 approach.

\paragraph{Verification.} We first tested a strategy of adding a ``perform OCR recognition first'' instruction in the prompt:

\begin{promptbox}[title={Prompt for OCR First Strategy}]
\begin{lstlisting}[style=promptstyle]
OCR first, then solve the problem.
\end{lstlisting}
\end{promptbox}

However, this strategy did not improve the results (Table~\ref{tab:ocr_verification}), possibly because the model focused its attention on OCR rather than reasoning. Therefore, we adopted the approach of directly inputting images. We use \texttt{gpt-5.2-1211-global} to perform error analysis on all samples where the model answered the text version correctly but the image version incorrectly. Errors are classified into information reading errors and reasoning errors using the following prompt:

\begin{promptbox}[title={Prompt for Error Classification}]
\begin{lstlisting}[style=promptstyle]
You are given:

1) **Text Problem**:  
`{text_problem}`

2) **Correct Text Solution** (guaranteed correct):  
`{text_solution}`

Now the same problem has been converted into an **equivalent image** (same content, just visual). A model answered based on the image and produced:

3) **Model Answer (from the image)**:  
`{model_answer}`

This model answer is **wrong**.

**Task:** Determine the most likely error type:

- **Information Reading Error**: The model misread, overlooked, or failed to extract key information from the image (e.g., missed a condition, misread a number/symbol, ignored part of the diagram/text).
- **Reasoning Error**: The model correctly extracted the relevant information, but made a logical/mathematical mistake in reasoning or computation.

**Instructions:**
1. Briefly list the key information that must be extracted from the problem to solve it.
2. Compare the **Model Answer** with the **Correct Text Solution** and infer whether the model likely failed at reading/extraction or at reasoning.
3. Output **only one label**: `Information Reading Error` or `Reasoning Error`.
4. Then give a **one-sentence justification**.
\end{lstlisting}
\end{promptbox}

\begin{table}[htb]
  \caption{Accuracy comparison and error analysis on MATH500 and its three image versions. ``Acc (OCR first)'' refers to performance when explicitly instructing the model to perform OCR before reasoning. Perception and reasoning error percentages are computed from samples where the model answered correctly on the text version but incorrectly on the image version.}
  \label{tab:ocr_verification}
  \centering
  \small
  \begin{tabular*}{\linewidth}{@{\extracolsep{\fill}}lcccc}
    \toprule
    \textbf{Dataset} & \textbf{Acc} & \makecell{\textbf{Acc}\\\textbf{(OCR first)}} & \makecell{\textbf{Perception}\\\textbf{Error}} & \makecell{\textbf{Reasoning}\\\textbf{Error}} \\
    \midrule
    MATH500 & 92.8 & - & - & - \\
    \midrule
    MATH500-v1 & 89.0 & 87.4 & 31.8\% (7/22) & 68.2\% (15/22) \\
    MATH500-v2 & 88.2 & 86.8 & 26.9\% (7/26) & 73.1\% (19/26) \\
    MATH500-v3 & 87.4 & 86.8 & 31.0\% (9/29) & 69.0\% (20/29) \\
    \bottomrule
  \end{tabular*}%
\end{table}

\section{Model Information}
\label{app:model_info}

Table~\ref{tab:model_arch} provides detailed architectural specifications for the three multimodal MoE models evaluated in this work.

\begin{table}[htb]
\caption{Model architectures. All expert counts are specified per layer.}
\label{tab:model_arch}
\centering
\small
\begin{tabular*}{\linewidth}{@{\extracolsep{\fill}}cccccc}
\toprule
\textbf{Model} & \textbf{Layers} & \makecell{\textbf{Routed}\\\textbf{Experts}} & \makecell{\textbf{Active Routed}\\\textbf{Experts}} & \makecell{\textbf{Shared}\\\textbf{Experts}} & \makecell{\textbf{Total Active}\\\textbf{Experts}} \\
\midrule
Kimi-VL & 27 & 64 & 6 & 2 & 8 \\
Qwen3-VL & 48 & 128 & 8 & -- & 8 \\
Llama4 & 48 & 16 & 1 & 1 & 2 \\
\bottomrule
\end{tabular*}%
\end{table}

\needspace{8\baselineskip}
\section{Evaluation Benchmarks}
\label{app:benchmarks}

\subsection{Benchmark Details}

For MATH500 and GPQA-Diamond, we render all problems as high-resolution images following the v2 approach described in \S\ref{sec:routing_divergence} to construct semantically equivalent visual inputs.

\begin{itemize}
\item \textbf{MATH500} is a curated subset of 500 problems from the MATH dataset.

\item \textbf{GPQA-Diamond} consists of 198 graduate-level multiple-choice questions in biology, physics, and chemistry. Given the limited number of biology questions (19), we select only the physics (86) and chemistry (93) subsets for evaluation.

\item \textbf{MathVerse} is a visual math benchmark spanning plane geometry, solid geometry, and function graphs, with six versions offering varying degrees of multimodal information. We adopt its vision-only version (788 test samples) to maximize visual dependency.

\item \textbf{MATH-Vision} is a collection of 3,040 mathematical problems with visual contexts sourced from real math competitions. We use the mini test set (304 samples) for evaluation.

\item \textbf{GSM8K-V} systematically transforms GSM8K text problems into purely visual multi-image format, comprising 1,319 samples (5,343 images) that require models to extract information from scene images. We use the implicit version to maximize visual dependency.
\end{itemize}

\subsection{MathVision Description Generation}
\label{app:mathvision_desc}

For MATH-Vision, we generate textual descriptions of diagrams to construct text references for domain expert identification. We use \texttt{gpt-5.2-1211-global} to generate image descriptions, excluding problem types where answers (such as counting) are directly evident in the description. The generation prompt follows the template shown below:

\begin{promptbox}[title={Prompt for MathVision Description Generation}]
\begin{lstlisting}[style=promptstyle]
Here is a diagram of a math problem, please describe the diagram in complete detail so that your description can be used to replace the diagram for math problem-solving.
\end{lstlisting}
\end{promptbox}
\paragraph{Example.} For the sample image shown in Figure~\ref{fig:mathvision_sample}, the generated description is:

\begin{figure}[h]
  \centering
  \includegraphics[width=0.8\linewidth]{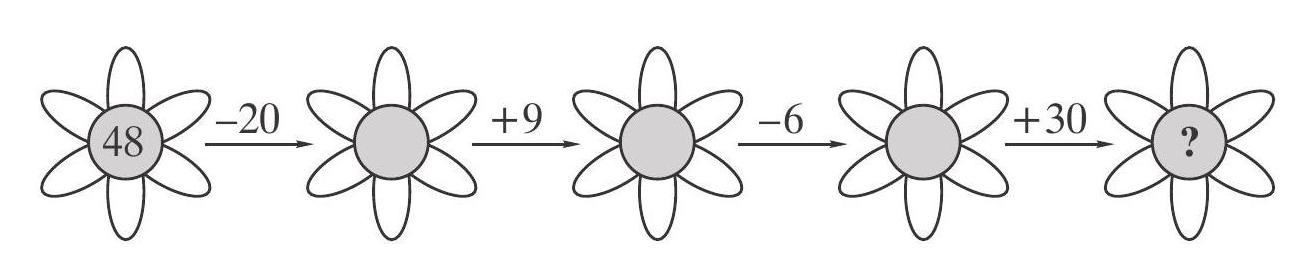}
  \caption{Sample image from MATH-Vision dataset.}
  \label{fig:mathvision_sample}
\end{figure}

\begin{promptbox}
\begin{lstlisting}[style=promptstyle]
A horizontal sequence of five identical flower shapes is shown, connected left-to-right by arrows.

- **Flower 1 (far left):** In the center circle is the number **48**.
- An arrow points from Flower 1 to **Flower 2** with the label **-20** above the arrow.
- **Flower 2:** Center circle is blank (no number shown).
- An arrow points from Flower 2 to **Flower 3** with the label **+9** above the arrow.
- **Flower 3:** Center circle is blank.
- An arrow points from Flower 3 to **Flower 4** with the label **-6** above the arrow.
- **Flower 4:** Center circle is blank.
- An arrow points from Flower 4 to **Flower 5 (far right)** with the label **+30** above the arrow.
- **Flower 5:** Center circle contains a **question mark (?)**.

Each flower has a round center and multiple petals (a simple daisy-like icon); only the center contents (48, blanks, and ?) matter for the math.
\end{lstlisting}
\end{promptbox}
These descriptions are concatenated with the original problem text to extract activation frequencies for domain expert identification.

\section{Expert Specialization Analysis}
\label{app:theory}

Expert specialization is an expected outcome of training in current mainstream MoE architectures. The standard load-balancing loss is:
\begin{equation}
L_{\text{balance}} = N_E \sum_{i=1}^{N_E} f_i \cdot p_i
\end{equation}
where $f_i$ is the activation frequency of expert $i$ and $p_i$ is its average routing probability.

In early MoE frameworks such as OpenMoE \cite{xue2024openmoe}, this loss is computed at the \textbf{micro-batch level}. Since tokens within a micro-batch often come from the same domain, the loss forces them to be distributed uniformly across all experts, preventing any expert from becoming domain-specific.

Modern architectures address this through two strategies. (a)~\textbf{Global-batch auxiliary loss} (e.g., Qwen3): computing $f_i$ over the entire global batch imposes a looser constraint, permitting local imbalance within a domain and explicitly encouraging specialization. (b)~\textbf{Auxiliary-loss-free routing} (e.g., DeepSeek-V3, Kimi-VL): dynamic bias terms guide load balancing without coupling it to the training gradient, granting experts maximum freedom to specialize. Both strategies have been shown to produce significant domain specialization \cite{qiu2025demons}.

\section{Implementation Details}
\label{app:eval_details}

Generation parameters are set to: temperature=0 (greedy decoding), zero-shot prompting, and maximum generation length of 8192 tokens, with no external tools enabled. All experiments are conducted on 16 A100 GPUs using vLLM v0.11.0. We wrap the MoE router in vLLM via EasySteer to intercept and modify routing logits during inference.

We observe that Qwen3-VL-30B-A3B-Instruct tends to generate excessively long chains of thought, likely due to long-CoT cold-start during its training phase. To mitigate this, we set repetition\_penalty=1.2 for all benchmarks except MATH500.

\end{document}